\documentclass[11pt]{article}

\usepackage[preprint]{acl}

\usepackage{times}
\usepackage{latexsym}

\usepackage[T1]{fontenc}

\usepackage[utf8]{inputenc}

\usepackage{microtype}

\usepackage{inconsolata}
\usepackage{float}

\usepackage{graphicx}

\usepackage{amsmath}
\usepackage{amssymb} 

\usepackage{placeins}
\usepackage{dsfont} 

\usepackage{listings}
\usepackage{booktabs}

\usepackage{multirow}
\usepackage[table]{xcolor}
\usepackage{booktabs}
\usepackage{placeins}

\usepackage{colortbl}
\definecolor{groupgray}{gray}{0.92}

\lstdefinestyle{promptstyle}{
  basicstyle=\ttfamily\footnotesize,
  columns=fullflexible,
  breaklines=true,
  breakatwhitespace=false,
  frame=single,
  numbers=left,
  numberstyle=\tiny,
  xleftmargin=2.2em,
  framexleftmargin=1.8em,
  showstringspaces=false,
  tabsize=2
}

\title{LeakageBench: Document-Level Leakage Risk for Redacting
Personally Identifiable Information in Document Images}

\author{
  Vishnu Prasad Vijaya Kumar$^{1,2}$ \quad
  Santhosh Venkatesh$^{2}$ \quad
  Ivan P.\ Yamshchikov$^{1}$ \\
  $^{1}$Center for Artificial Intelligence and Robotics (CAIRO), \\
  Technical University of Applied Sciences W{\"u}rzburg-Schweinfurt (THWS), \\
  W{\"u}rzburg, Germany \\
  $^{2}$DataX, Frankfurt am Main, Germany \\
  {\small
  \href{mailto:vishnuprasad.vijayakumar@thws.de}
       {\nolinkurl{vishnuprasad.vijayakumar@thws.de}}
  \quad
  \href{mailto:santhosh.venkatesh@datax.me}
       {\nolinkurl{santhosh.venkatesh@datax.me}}
  } \\
  {\small
  \href{mailto:ivan.yamshchikov@thws.de}
       {\nolinkurl{ivan.yamshchikov@thws.de}}
  }
}

\begin{document}
\maketitle

\begin{abstract}
Real-world personally identifiable information (PII) redaction often
operates on document images---scans, screenshots, and PDF
renderings---where OCR errors, layout structure, and visual noise
determine whether sensitive information is actually removed. Existing
PII benchmarks are mostly text-centric and do not measure
document-level redaction risk: a page remains unsafe if even one
identifier is missed. We introduce LeakageBench, a challenge set of
500 document images with 11,954 GDPR-aligned PII annotations spanning
direct identifiers, linkage keys, and contextual re-identification
surfaces. We evaluate generic OCR pipelines, commercial and
task-adapted OCR-dependent detectors, and OCR-free vision-language
models using entity-level F1, group-wise leakage, and document-level
leakage metrics. Code Interpreter raises GPT-5.5 localization F1 from
0.090 to 0.249, but critical page-level leakage remains 0.968. These
results show that stronger detection and tool assistance improve
localization without making most pages safe for release. LeakageBench
provides a diagnostic benchmark for high-recall, spatially grounded
PII redaction in document images.
\end{abstract}

\section{Introduction}

Organizations routinely redact personally identifiable information
(PII) from scanned and rendered business documents. OCR errors,
complex layouts, repeated identifiers, and visual noise make this more than a clean-text tagging problem. Most PII benchmarks measure
span-level precision, recall, and F1 on text, while document-understanding benchmarks typically target selected key fields. Neither directly measures release safety: a page remains unsafe when any in-scope identifier remains visible. Evaluation must therefore capture both entity-level localization quality and residual page-level leakage.

Redaction presents an asymmetric operational trade-off. False
positives unnecessarily obscure useful document content, whereas a
single false negative can make a page unsafe to release. We therefore
treat document-level leakage as a recall-oriented complement to, rather than a replacement for, localization and typing F1, which also penalize over-redaction through unmatched predictions.

We introduce \textbf{LeakageBench}, a challenge set of \textbf{500}
document-page images containing \textbf{11,954} localized PII
annotations. Its GDPR-aligned schema distinguishes \emph{direct identifiers}, \emph{linkage keys}, and \emph{contextual re-identification surfaces}. Using a shared image-space interface,
LeakageBench exposes a stark deployment gap: even the strongest evaluated configuration leaks critical Direct+Linkage PII on 96.8\% of applicable pages. Our diagnostics attribute these failures to
schema coverage, OCR recoverability, detection, interface validity, and spatial grounding.

Our contributions are: (1) a controlled, densely annotated document-image redaction challenge set; (2) a privacy-oriented schema covering direct, linkage, and contextual identifiers; (3) release-oriented document-level leakage metrics reported alongside entity-level F1; and (4) a unified diagnostic evaluation identifying why OCR-dependent and OCR-free systems continue to leak PII.

\section{Related Work}

\paragraph{Text-based PII detection and anonymization.}
Text-based benchmarks cover legal documents, regulatory categories,
student writing, and clinical narratives
\citep{pilan-etal-2022-text,Gambarelli_2023,mo2025datasir,
pii-detection-removal-from-educational-data,STUBBS2015S11}.
They establish methods for span-level PII detection and anonymization,
but exclude document-image artifacts, spatial localization, and
page-level release safety.

\paragraph{Document-image understanding.}
FUNSD, SROIE, CORD, and DocVQA evaluate form understanding, receipt
extraction, OCR, and visually grounded question answering
\citep{jaume2019funsd,DBLP:journals/corr/abs-2103-10213,
park2019cord,DBLP:journals/corr/abs-2007-00398}.
BuDDIE broadens coverage to multiple business-document tasks and
includes person- and address-related fields
\citep{wang-etal-2025-buddie}. However, these benchmarks target
task-specific extraction rather than exhaustive PII localization and
residual redaction leakage.

\paragraph{OCR-based and OCR-free document models.}
OCR-based models such as LayoutLM and LayoutLMv3 combine recognized text with two-dimensional layout
\citep{DBLP:journals/corr/abs-1912-13318,
huang2022layoutlmv3pretrainingdocumentai}, whereas OCR-free models such as Donut generate structured outputs directly from document images \citep{kim2022donut}. For redaction, the former can propagate OCR and text-to-box alignment errors, while the latter must jointly learn visual reading and spatial localization. LeakageBench evaluates both through a common image-space prediction interface.

\paragraph{Leakage-oriented privacy evaluation.}
PRvL evaluates LLM-based PII redaction and residual exposure
\citep{garza2025prvlquantifyingcapabilitiesrisks}; ProPILE studies
privacy leakage through targeted probing
\citep{kim2023propileprobingprivacyleakage}; and RedactBuster shows that redacted documents can remain vulnerable to downstream inference
\citep{beltrame2024redactbusterentitytyperecognition}.
These works motivate exposure-oriented evaluation, but do not provide a localized document-image benchmark with a page-level release criterion.

\section{Task and Evaluation Protocol}
\label{sec:task}

\subsection{Task definition}
\label{sec:task_def}
We evaluate \emph{PII redaction on document images}. 
Given a page image---a scan, screenshot, or PDF rendering---a system must identify every visible region containing personally identifiable information, assign each region a schema type, and optionally extract its text value. Each gold instance is represented by an image-space bounding box, a fine-grained PII type, a risk group, and an optional verbatim value string.

The task is redaction-oriented rather than text-only: predictions must be spatially grounded so that the sensitive region can be masked on the image. This makes recall the primary safety concern. A page remains unsafe if even one in-scope identifier is missed, regardless of average entity-level performance. Inference is page-local: each rendered page is processed independently, and prediction–gold matching is performed only within that page. No cross-page context is provided. Accordingly, DocLeak is computed over document pages; evaluating multi-page contextual inference remains future work.

\subsection{Prediction format}
\label{sec:pred_format}
A system outputs a set of predicted instances
\[
  \hat{y}=\{(\hat{b}_j,\hat{t}_j,\hat{v}_j,\hat{s}_j)\}_{j=1}^{M},
\]
where $\hat{b}_j=(x_{\min},y_{\min},x_{\max},y_{\max})$ is an axis-aligned bounding box in pixels, $\hat{t}_j$ is a PII type, $\hat{v}_j$ is an optional extracted value string, and $\hat{s}_j\in[0,1]$ is an optional confidence score.
This unified interface supports both OCR-assisted pipelines, where boxes and values are derived from OCR spans, and OCR-free vision-language models.

\subsection{Instance matching}

We score predictions using one-to-one matching within each page. A predicted
box $\hat{b}$ is eligible to match a ground-truth box $b$ when
$\operatorname{IoU}(\hat{b},b)\geq\tau$. We enumerate all eligible pairs and
greedily select them in descending IoU order, ensuring that each prediction and
ground-truth instance is used at most once. A selected pair counts as a
localization match regardless of type and as a typed match only when the
predicted and ground-truth PII types agree. When values are available, value
correctness is evaluated only for matched pairs.

Headline DocLeak uses type-aware matching: a gold instance is counted
as protected only when an IoU-eligible prediction has the same PII
type. This conservative criterion is used in
Tables~\ref{tab:main_results}--\ref{tab:supported_schema}. Diagnostic
attribution additionally uses type-agnostic matching because a wrongly
typed but correctly localized box would still mask the identifier.

\subsection{Metrics}
\label{sec:metrics}

We report entity-level F1 and document-level leakage. The former measures
average detection quality; the latter measures whether any PII remains
unmatched on a page.

\paragraph{Entity-level performance.}
$\mathrm{F1}_{\mathrm{loc}}$ is page-macro localization F1 using
type-agnostic matching. $\mathrm{F1}_{\mathrm{type}}$ requires both
$\operatorname{IoU}\geq\tau$ and the correct PII type, and is macro-averaged
over ground-truth types. Unless stated otherwise, $\tau=0.75$.

\paragraph{Document-level leakage.}
A page leaks if at least one in-scope ground-truth instance is unmatched:
\begin{equation}
    \mathrm{DocLeak}
    =
    \frac{1}{|\mathcal{D}|}
    \sum_{d\in\mathcal{D}}
    \mathbb{1}[d\text{ leaks}],
\end{equation}
where $\mathcal{D}$ contains pages with at least one in-scope identifier.
We report $\mathrm{DocLeak}_{\mathrm{all}}$ over all PII and
$\mathrm{DocLeak}_{\mathrm{crit}}$ over Direct and Linkage identifiers.

Headline DocLeak uses type-aware matching. The error analyses additionally
use type-agnostic matching to measure whether an identifier would be
spatially masked despite an incorrect type. DocLeak penalizes unmatched gold instances but does not penalize extra predictions. We therefore report it jointly with F1$_{\mathrm{loc}}$ and F1$_{\mathrm{type}}$, whose precision terms count unmatched predictions as false positives. One-to-one IoU matching prevents a single prediction from receiving credit for
multiple gold instances.

\section{Dataset: Collection and Composition}
\label{sec:dataset}

\subsection{Sources and generation process}
\label{sec:sources}
\textbf{LeakageBench} contains \textbf{500} document-page images sampled from public operational-style business documents. We position the dataset as a challenge set for stress-testing image-based PII redaction under heterogeneous layouts, dense tables, scanned correspondence, mixed acquisition channels, and visual degradation. Rather than claiming to represent all deployment settings, LeakageBench covers a focused slice of business-document redaction workflows where identifiers appear across forms, invoices, emails, letters, and free-form correspondence.

\paragraph{Source A: OCR-IDL (public archive documents).}
We sample \textbf{103} documents (113) pages from OCR-IDL \citep{biten2022ocridlocrannotationsindustry}, which provides standardized OCR outputs for documents from the UCSF Industry Documents Library (IDL) \footnote{\url{https://industrydocuments.ucsf.edu/}}, a large digital archive of industry documents hosted by UCSF.
OCR-IDL is valuable for our setting because it includes heterogeneous, historically scanned/archived pages with substantial layout and rendering variation.

\paragraph{Source B: VRDU Ad-Buy Forms (structured, table-heavy invoices/receipts).}
We sample \textbf{107} documents (222) pages from the VRDU Ad-buy Forms corpus released by Google Research \citep{Wang_2023}.
These documents are public Federal Communications Commission (FCC) \footnote{\url{https://publicfiles.fcc.gov/}} filings and are dominated by structured invoices/receipts with multi-column layouts and repeated/nested table structures, providing a concentrated stress-test for localization within dense tabular regions.

\paragraph{Source C: Unstructured operational-style documents (free-form correspondence).}
We sample the remaining \textbf{81} documents (165) pages from FCC public filings, selecting free-form pages that resemble internal operational documents: email threads and attachments, letters and memos, scanned correspondence, and mixed-format ``file'' pages.
Compared to invoice- or form-heavy sources, these pages exhibit less templated structure and place identifiers in narrative body text, signatures, headers/footers, and embedded attachments.
This source therefore stresses redaction systems under realistic cross-field co-reference (the same individual referenced across multiple regions) and non-field PII placement.

\subsection{Identifiability-first scope}
\label{sec:scope}

Privacy regulation is triggered by \emph{identifiability}, not by whether a string resembles a name.
Under the GDPR, personal data includes identifiers such as names, identification numbers, and location data (Art.~4(1)) \citep{gdpr}.
Recital~26 emphasizes that identifiability must consider \emph{all means reasonably likely to be used}, including \emph{singling out} and linkage by other parties \citep{gdpr}.
Similarly, the CPRA/CCPA defines personal information as data that could reasonably be linked (directly or indirectly) to a consumer or household \citep{cpra_1798_140},
and explicitly recognizes that personal information can exist in physical formats, including paper documents and printed images \citep{cpra_1798_140}.
In U.S. standards, PII is likewise defined to include information that is \emph{linked or linkable} to an individual \citep{nist_sp800_122}.

Consequently, a redaction benchmark must treat as in-scope not only direct identifiers, such as names and contacts, but also \emph{linkage keys}
(policy numbers, case IDs, customer IDs) that resolve to a person through common workflows.

\subsection{PII schema: re-identification surfaces}
\label{sec:schema}
We annotate fine-grained labels for auditing, but report results using three privacy-aligned groups that reflect how identifiability arises in operational document workflows: (i) direct identification, (ii) lookup/linkage, and (iii) contextual inference.
Table~\ref{tab:pii_schema_groups} summarizes the grouping used for reporting and analysis and lists the fine-grained labels covered under each group.

\begin{table*}[t]
\centering
\small
\setlength{\tabcolsep}{6pt}
\renewcommand{\arraystretch}{1.12}
\begin{tabular}{p{0.16\textwidth} p{0.54\textwidth} p{0.22\textwidth}}
\hline
\textbf{Group} & \textbf{Fine-grained labels (annotated)} & \textbf{Basis (examples)} \\
\hline
Direct identifiers &
\textsc{Person Name, Address, Email Address, Phone Number} &
GDPR; CPRA/CCPA \citep{gdpr,cpra_1798_140} \\
\hline
Linkage keys &
\textsc{Customer ID, Employee ID, Policy No., ID Card No., Case/File/Reference No., Contract/Invoice/Cheque No., Registration No., Call Sign, DA Case Number} &
NIST; HIPAA; CPRA/CCPA \citep{nist_sp800_122,hipaa_164_514,cpra_1798_140} \\
\hline
Contextual identifiers &
\textsc{Date, Location, Organization Name, Business ID} &
GDPR Recital~26; NIST \citep{gdpr,nist_sp800_122} \\
\hline
\end{tabular}
\caption{Privacy-aligned schema groups and supporting regulatory anchors. We annotate fine-grained labels and additionally report aggregated results by group.}
\label{tab:pii_schema_groups}
\end{table*}

\paragraph{Rationale.}
\textbf{Direct identifiers} explicitly reveal identity or contact information \citep{gdpr,cpra_1798_140}.
\textbf{Linkage keys} identify via lookup: they may appear innocuous on the page but act as stable keys into organizational systems or registries, consistent with standards that treat unique identifying numbers and codes as identifying \citep{nist_sp800_122,hipaa_164_514} and with CPRA/CCPA ``unique identifiers'' \citep{cpra_1798_140}.
Finally, we annotate \textbf{contextual identifiers} (e.g., dates and entity fields) because they increase identifiability through co-occurrence and linkage in real documents \citep{gdpr,nist_sp800_122}.
We do not assume each contextual field is personal data in isolation; rather, it captures \emph{re-identification surfaces} that amplify risk when combined with direct identifiers or linkage keys.

\subsection{Dataset statistics and diagnostic views}
\label{sec:stats}
\begin{figure*}[t]
\centering
\begin{minipage}{0.49\textwidth}
  \centering
  \includegraphics[width=\linewidth]{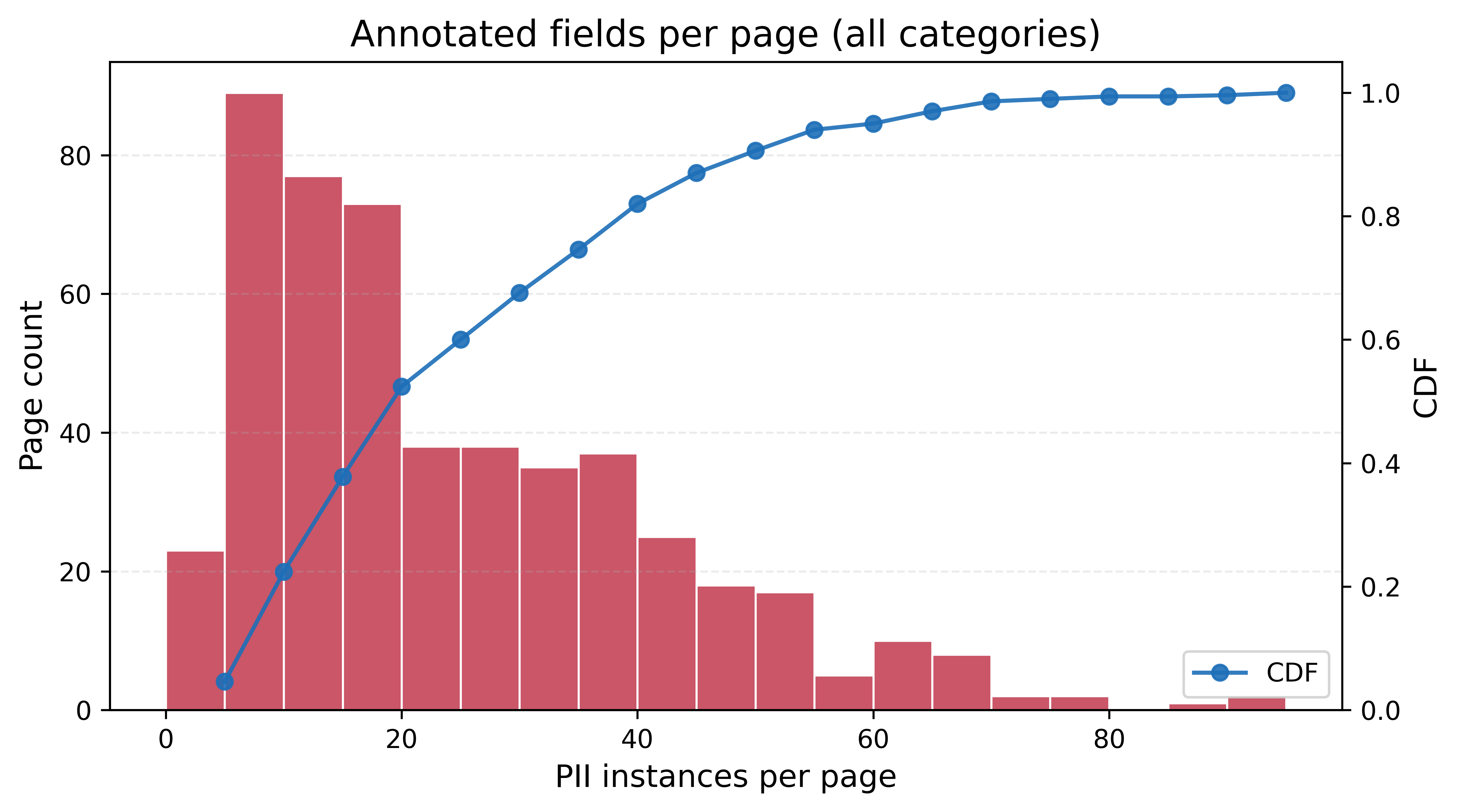}
\end{minipage}\hfill
\begin{minipage}{0.49\textwidth}
  \centering
  \includegraphics[width=\linewidth]{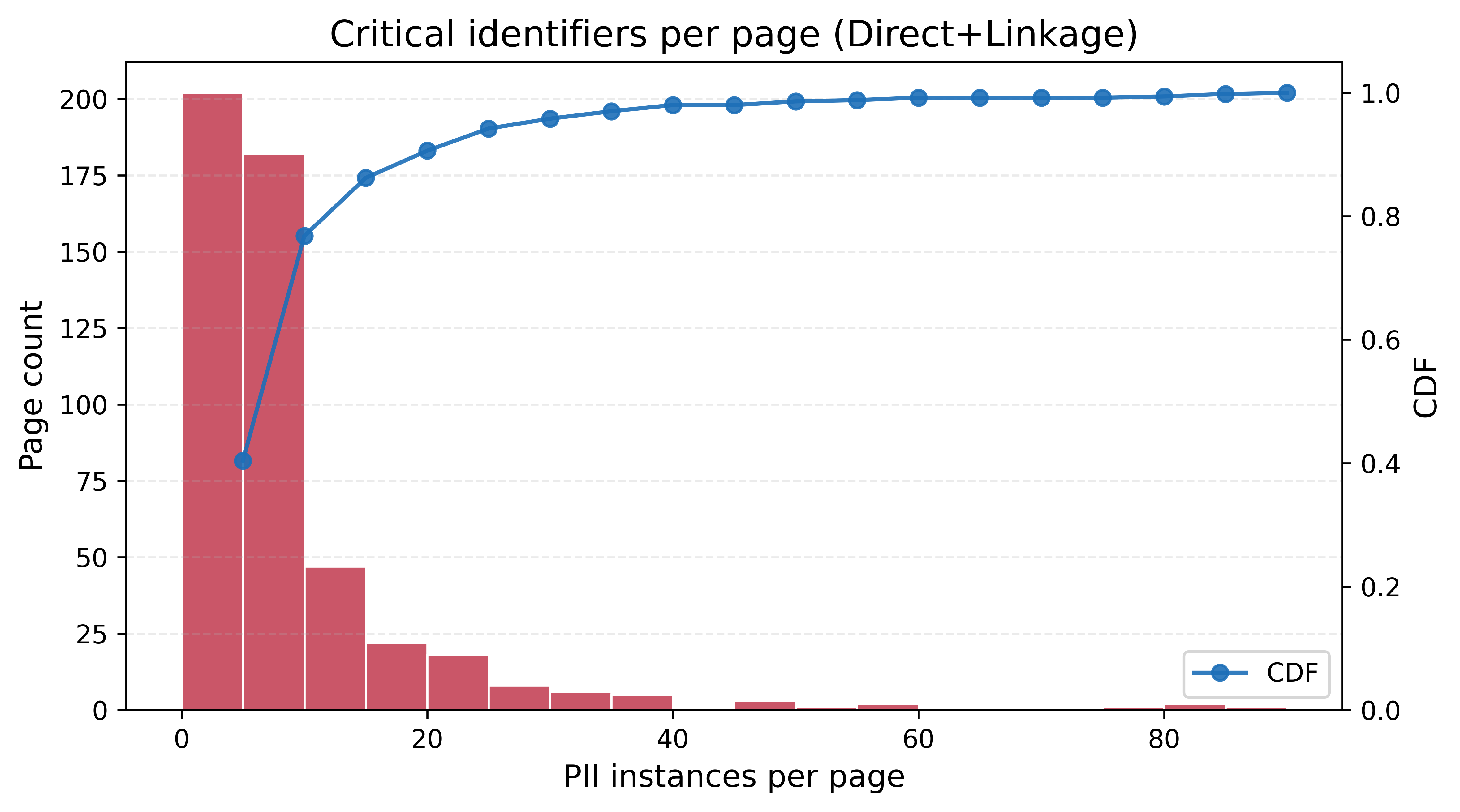}
\end{minipage}
\caption{Per-page PII density in LeakageBench (histogram with CDF overlay). Left: all annotated PII. Right: safety-critical identifiers (Direct+Linkage).}
\label{fig:pii_hist}
\end{figure*}

\begin{figure*}[t]
\centering
\begin{minipage}{0.40\textwidth}
  \centering
  \includegraphics[width=\linewidth]{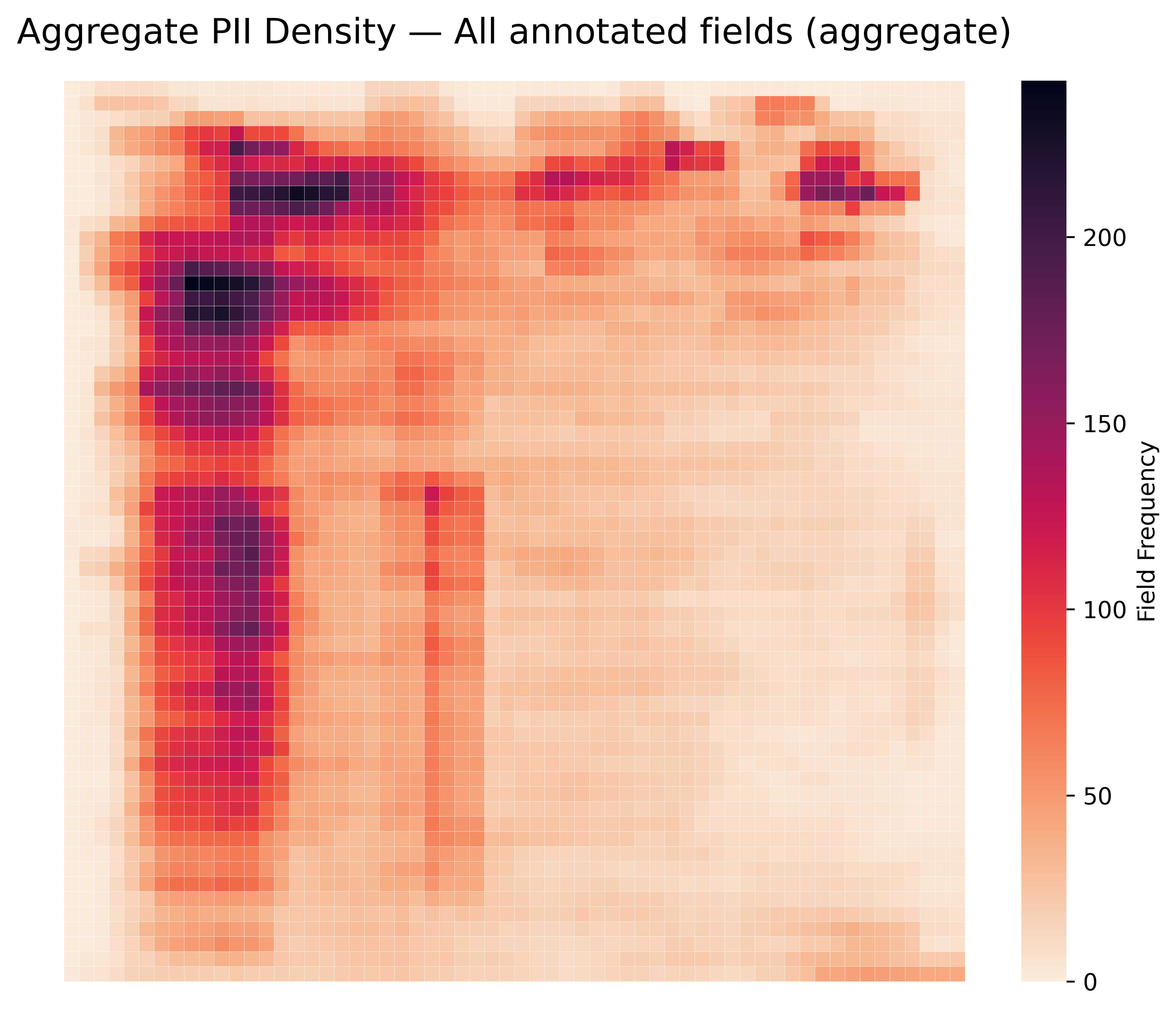}
\end{minipage}\hfill
\begin{minipage}{0.40\textwidth}
  \centering
  \includegraphics[width=\linewidth]{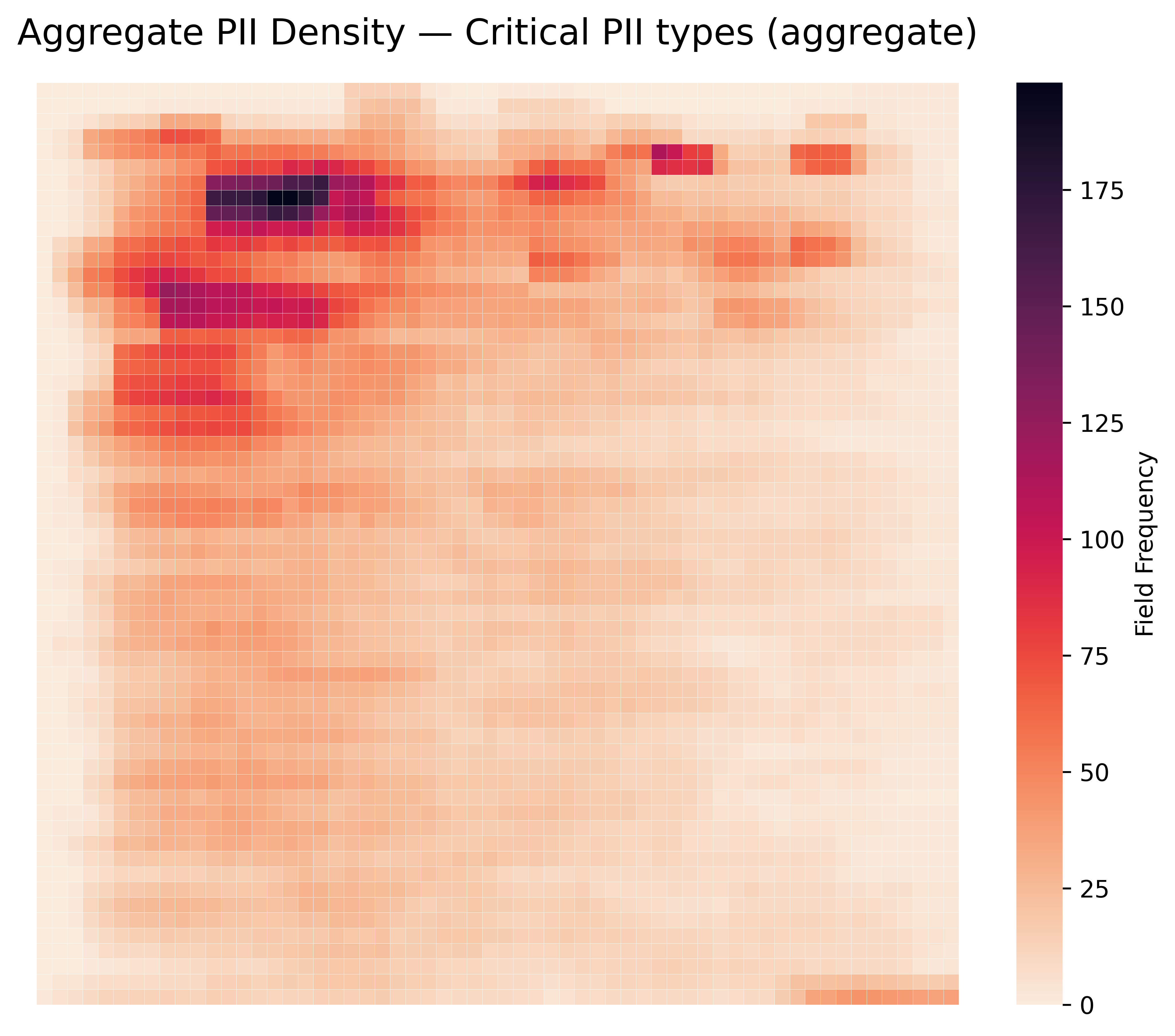}
\end{minipage}
\caption{Spatial PII density aggregated over pages. Left: all annotated fields. Right: critical identifiers (Direct+Linkage).}
\label{fig:pii_heatmap}
\end{figure*}

Table~\ref{tab:dataset_stats} reports composition and label volume for \textbf{LeakageBench} (released) and for each source, including document and page counts and PII group totals.
Redaction safety is determined by document-level worst cases. We therefore provide diagnostic views that expose the benchmark's risk profile: (i) the identifiers-per-page distribution (Fig.~\ref{fig:pii_hist}) and (ii) the spatial concentration of PII on the page (Fig.~\ref{fig:pii_heatmap}).

\paragraph{Identifiers-per-page distribution.}
Figure~\ref{fig:pii_hist} shows the distribution of PII instances per page.
This distribution is critical for safety: a page leaks if any identifier is missed, so pages with many identifiers place disproportionate pressure on redaction systems.
Across the collected pool, we annotate \textbf{11{,}954} PII instances over \textbf{500} pages (\textbf{23.90} per page on average), including \textbf{4{,}315} critical identifiers from the Direct and Linkage categories (\textbf{8.63} per page on average).
We therefore report document-level leakage alongside entity-level scores (Sec.~\ref{sec:metrics}).

\paragraph{Spatial PII heatmap.}
Figure~\ref{fig:pii_heatmap} visualizes where identifiers appear on the page.
We partition each image into a regular grid of non-overlapping 14×14-pixel patches (stride = 14), yielding 3,600 patches per page (60 × 60).
For each patch, we record which PII categories overlap its area and aggregate the results across all pages.
The resulting heatmap distinguishes benchmarks where PII concentrates in predictable regions-such as headers or signatures-from those where it is distributed throughout tables and body text.
We report both an ``all PII'' heatmap and a ``critical PII'' heatmap restricted to Direct+Linkage identifiers.

\begin{table*}[t]
\centering
\small
\setlength{\tabcolsep}{6pt}
\renewcommand{\arraystretch}{1.12}
\begin{tabular}{l r r r r r r r}
\hline
\textbf{Source} & \textbf{\#Docs} & \textbf{\#Pages} &
\textbf{Direct} & \textbf{Linkage} & \textbf{Contextual} & \textbf{Other} &
\textbf{Critical/Page} \\
\hline
Source A (OCR-IDL) & 103 & 113  & 1{,}143 & 109 & 445 & 17 & 11.08 \\
Source B (VRDU Ad-Buy) & 107 & 222  & 673 & 221 & 6{,}245 & 192 & 4.03 \\
Source C (FCC free-form) & 81 & 165  & 1{,}963 & 206 & 710 & 30 & 13.15 \\
\hline
\textbf{Total} & 291 & 500  & 3{,}779 & 536 & 7{,}400 & 239 & 8.63 \\
\hline
\end{tabular}
\caption{Source composition and label volume. \textbf{Other} aggregates non-group labels (e.g., Signature, Agency Code/Ref, Advertiser Ref). \textbf{Critical/Page} is (Direct+Linkage)/Pages.}
\label{tab:dataset_stats}
\end{table*}

\paragraph{Document types.}
LeakageBench is dominated by operational business documents: invoices, emails, forms, letters/correspondence, and file-like pages. 
Full page-type composition by source is reported in Appendix~\ref{app:doc_types}.

\subsection{Annotation procedure}
\label{sec:annotation}

All annotations were produced by the first two authors. The 500 page images were randomly split equally and annotated independently following the guidelines in Appendix~\ref{app:annotation_instructions}. Annotators labeled entity instances using tight bounding boxes around visible PII spans and transcribed \texttt{value} strings verbatim (no normalization); multi-line items use the delimiter ``\texttt{\textbackslash n}''. We annotate fine-grained entity types and additionally report privacy-aligned group aggregates (Direct/Linkage/Contextual) for analysis.

\subsection{Annotation quality control}
\label{sec:annotation_quality}
To assess annotation consistency, each annotator contributed a 10\% stratified sample of their assigned pages for cross-annotation by the other annotator (\textbf{50 images / pages total}). Disagreements were resolved by joint adjudication to form a single gold set.

Inter-annotator agreement is measured with Cohen’s $\kappa$ over entity instances after alignment.
We align annotations via one-to-one bipartite matching using IoU $\ge \tau$ with $\tau{=}0.75$.
A matched pair is counted as agreement only if both boxes share the same fine-grained type; spatial overlaps with a type mismatch are counted as disagreements.
Any annotation with no IoU$\ge0.75$ counterpart from the other annotator is treated as a miss (disagreement).
Under this protocol, we obtain $\kappa = 0.979$ on the cross-annotated subset.

\section{Experimental Setup}
\label{sec:exp}

We evaluate off-the-shelf systems under one unified image-level redaction protocol. 
The baselines cover two architectures. 
First, \textbf{OCR-dependent systems} read the page with OCR, detect PII in the recognized text, and project detections back to image coordinates using OCR word boxes. 
Within this family, we compare a generic redaction pipeline against stronger commercial and task-adapted privacy detectors. 
Second, \textbf{OCR-free vision-language models} receive the page image directly and predict PII regions without an explicit OCR stage. 
All systems are converted to the same prediction format (Sec.~\ref{sec:pred_format}) and evaluated with entity-level localization, localization+typing, document-level leakage, schema coverage, and supported-type leakage (Sec.~\ref{sec:metrics}). 
No system is trained or tuned on LeakageBench. 

\subsection{OCR-dependent baselines}
\label{sec:exp_ocr}
All OCR-dependent baselines use the same image-to-box pipeline: OCR extracts text and word-level boxes, a privacy detector predicts typed PII spans over the OCR text, and each span is mapped back to the page by merging the corresponding word boxes.

\paragraph{Generic OCR-dependent redaction.}
We use Microsoft Presidio Image Redactor \cite{presidio} as the generic OCR-dependent baseline. 
Presidio applies OCR, detects PII spans with Presidio Analyzer, and returns image-space redaction boxes through OCR word alignment. 
We use Tesseract \cite{smith2007tesseract} for this baseline and map Presidio entity types to the LeakageBench schema with a fixed mapping.

\paragraph{Commercial and task-adapted OCR-dependent detectors.}
We evaluate stronger privacy-oriented detectors on top of Amazon Textract OCR \cite{aws_textract}: GLiNER-base \cite{zaratiana-etal-2024-gliner}, Presidio-best (GLiNER restricted), GLiNER multi-PII \cite{gliner_multi_pii}, NVIDIA GLiNER PII \cite{nvidia_gliner_pii}, GLiNER2 \cite{zaratiana-etal-2025-gliner2}, Google DLP \cite{google_dlp}, Amazon Comprehend PII \cite{aws_comprehend_pii} and an OpenAI Privacy Filter \cite{openai_privacy_filter} pipeline. 
These systems test whether stronger text-side PII detectors are sufficient once document-image constraints---OCR errors, dense layouts, repeated identifiers, span fragmentation, and box reconstruction---are introduced.
\paragraph{Schema alignment.}
All OCR-dependent systems are scored against the LeakageBench schema. 
Unsupported entity types count as missed detections under full-schema evaluation. 
We therefore report schema coverage and supported-type leakage to separate taxonomy limitations from empirical detection and localization failures.

Appendix~\ref{app:baseline_details} summarizes the OCR engines, detector settings, schema mappings, and conversion rules for each OCR-dependent baseline.

\subsection{OCR-free vision-language baselines}
\label{sec:exp_ocrfree}

\paragraph{OCR-free vision-language localization.}
We evaluate \textbf{Qwen3-VL-32B} \cite{bai2025qwen3vltechnicalreport} and \textbf{InternVL3-38B} \cite{zhu2025internvl3exploringadvancedtraining} as OCR-free vision-language baselines because they support explicit 2D grounding from image input. 
Unlike OCR-dependent systems, these models receive the page image directly and are prompted to return PII types, values, confidence scores, and bounding boxes. 
Predicted boxes are converted to image coordinates and evaluated under the same localization-and-typing protocol as all other systems. 
Coordinate conventions and parsing rules are summarized in Appendix~\ref{app:baseline_details}; full prompt templates are provided in Appendix~\ref{app:prompts}.

\paragraph{Closed OCR-free models.}
We additionally evaluate GPT-5.4 using single-pass vision and GPT-5.5 in two
configurations: single-pass vision and Code Interpreter-assisted inspection.
All configurations use the same PII taxonomy and semantic instructions and
return structured predictions with integer bounding boxes in a normalized
$0$--$999$ coordinate system. For GPT-5.5, we use high image detail and medium
reasoning effort; in the tool-assisted configuration, Code Interpreter is
required to inspect the page and may crop or zoom regions before returning the
final prediction. All outputs are processed using the same parsing, coordinate
conversion, and evaluation pipeline as the other OCR-free models.


\paragraph{Prompt-driven schema coverage.}
For OCR-free systems, the full LeakageBench schema is supplied in the prompt. 
Their schema coverage is therefore $1.0$ by configuration: the model is allowed to emit every benchmark type, but this does not imply reliable detection. 
Empirical failures are captured by entity-level F1, document-level leakage, group leakage, and output-validity rates.

\paragraph{Why not LayoutLM-style trained baselines.}
LeakageBench is evaluated as an off-the-shelf challenge set: systems must run without training on the benchmark and must return PII types with image-space boxes. LayoutLM-style document NER models typically require supervised fine-tuning, task-specific heads, OCR-token alignment, and a train/dev/test split.  We therefore leave layout-aware trained models to a future trained-baseline track.

\section{Results}
\label{sec:results}
Table~\ref{tab:main_results} reports full-schema performance at IoU $\tau=0.75$ under the unified prediction format. To assess sensitivity to box tightness, Appendix~\ref{app:iou_sweep} reports results at $\tau \in \{0.50,0.75,0.95\}$.
We report localization F1, localization+typing F1, document-level leakage over all identifiers, and leakage over safety-critical Direct+Linkage identifiers. 
Unless stated otherwise, unsupported schema types, invalid outputs, and out-of-bounds boxes are counted as missed detections.

For systems with restricted taxonomies, we additionally report supported-type leakage in Table~\ref{tab:supported_schema}, where gold labels are filtered to the entity types each system can emit. 
Schema coverage and full-schema group-wise leakage are reported in Appendix~\ref{app:schema_coverage} and Appendix~\ref{app:group_leakage}.
Valid denotes the fraction of pages producing parseable, schema-conformant outputs. Malformed or unusable outputs are scored as empty prediction sets; individual invalid boxes are rejected and receive no credit.
\begin{table*}[t]
\centering
\small
\setlength{\tabcolsep}{5.5pt}
\renewcommand{\arraystretch}{1.12}

\begin{tabular}{l l r r r r r}
\hline
\textbf{System} &
\textbf{OCR} &
\textbf{F1$_{\mathrm{loc}}\uparrow$} &
\textbf{F1$_{\mathrm{type}}\uparrow$} &
\textbf{DocLeak$_{\mathrm{all}}\downarrow$} &
\textbf{DocLeak$_{\mathrm{crit}}\downarrow$} &
\textbf{Valid$\uparrow$} \\
\hline

\rowcolor{groupgray}
\multicolumn{7}{l}{
\textbf{Generic baseline}
\hspace{0.4em}\textit{(Tesseract OCR)}
} \\
Presidio
& Tesseract
& 0.137
& 0.048
& 0.994
& 0.990
& 0.996 \\

\rowcolor{groupgray}
\multicolumn{7}{l}{
\textbf{Shared AWS Textract OCR}
\hspace{0.4em}\textit{(detector varies; OCR is fixed)}
} \\
Presidio-best (GLiNER restr.)
& Textract
& 0.248
& 0.064
& 0.996
& 0.986
& \textbf{1.000} \\
GLiNER-base
& Textract
& 0.216
& 0.063
& \textbf{0.988}
& 0.978
& \textbf{1.000} \\
GLiNER-multi-PII
& Textract
& 0.245
& 0.091
& 0.990
& 0.972
& \textbf{1.000} \\
NVIDIA GLiNER PII
& Textract
& 0.288
& 0.109
& 0.992
& 0.986
& \textbf{1.000} \\
GLiNER2
& Textract
& 0.156
& 0.065
& 1.000
& 0.994
& 0.998 \\
Amazon Comprehend PII
& Textract
& \textbf{0.304}
& 0.085
& 0.992
& \textbf{0.968}
& \textbf{1.000} \\
OpenAI Privacy Filter
& Textract
& 0.164
& 0.055
& 0.996
& 0.990
& 0.806 \\

\rowcolor{groupgray}
\multicolumn{7}{l}{
\textbf{Vendor-managed OCR}
} \\
Google DLP
& Vendor
& 0.076
& 0.013
& 1.000
& 1.000
& \textbf{1.000} \\

\rowcolor{groupgray}
\multicolumn{7}{l}{
\textbf{OCR-free vision-language models}
\hspace{0.4em}\textit{(no separate OCR stage)}
} \\
Qwen3-VL-32B
& ---
& 0.073
& 0.046
& 1.000
& 0.998
& 0.666 \\
InternVL3-38B
& ---
& 0.012
& 0.019
& 1.000
& 1.000
& 0.540 \\
GPT-5.4
& ---
& 0.044
& 0.031
& 1.000
& 1.000
& \textbf{1.000} \\
GPT-5.5
& ---
& 0.090
& 0.050
& 1.000
& 0.990
& \textbf{1.000} \\
GPT-5.5 + Code Interpreter
& ---
& 0.249
& \textbf{0.119}
& 0.990
& \textbf{0.968}
& \textbf{1.000} \\

\hline
\end{tabular}

\caption{
Full-schema results on \textbf{LeakageBench} at IoU $\tau=0.75$.
Textract systems share identical OCR, isolating detector differences.
F1$_{\mathrm{loc}}$ is page-macro type-agnostic localization F1;
F1$_{\mathrm{type}}$ also requires the correct type.
DocLeak$_{\mathrm{all}}$ and DocLeak$_{\mathrm{crit}}$ are the fractions of pages leaking any identifier and any Direct+Linkage identifier, respectively. Unsupported types and invalid outputs count as misses; Table~\ref{tab:supported_schema} controls for schema coverage. Valid denotes parseable, schema-conformant page outputs. Bold marks the best result, including ties.
}
\label{tab:main_results}
\end{table*}

\begin{table*}[t]
\centering
\small
\setlength{\tabcolsep}{3.6pt}
\renewcommand{\arraystretch}{1.10}

\begin{tabular}{l l r r r r r r}
\hline
\textbf{System} &
\textbf{OCR} &
\textbf{Coverage$\uparrow$} &
\textbf{Unsup.\ pages$\downarrow$} &
\textbf{Direct$^{\mathrm{sup}}\downarrow$} &
\textbf{Linkage$^{\mathrm{sup}}\downarrow$} &
\textbf{Contextual$^{\mathrm{sup}}\downarrow$} &
\textbf{All$^{\mathrm{sup}}\downarrow$} \\
\hline

\rowcolor{groupgray}
\multicolumn{8}{l}{
\textbf{Restricted schema coverage}
\hspace{0.4em}\textit{(unsupported gold types are excluded)}
} \\
Presidio
& Tesseract
& 0.208
& 0.900
& 0.973
& --
& 0.923
& 0.990 \\
Presidio-best
& Textract
& 0.208
& 0.900
& 0.967
& --
& 0.942
& 0.992 \\
GLiNER2
& Textract
& \textbf{0.958}
& \textbf{0.094}
& 0.971
& 0.985
& 0.963
& 1.000 \\
Google DLP
& Vendor
& 0.250
& 0.762
& 1.000
& --
& 0.990
& 0.998 \\
Amazon Comprehend PII
& Textract
& 0.208
& 0.900
& 0.908
& --
& 0.873
& 0.968 \\
OpenAI Privacy Filter
& Textract
& 0.250
& 0.850
& 0.971
& 0.994
& 0.991
& 0.990 \\

\rowcolor{groupgray}
\multicolumn{8}{l}{
\textbf{Full schema coverage with shared AWS Textract OCR}
\hspace{0.4em}\textit{(detector varies; OCR and schema are fixed)}
} \\
GLiNER-base
& Textract
& 1.000
& 0.000
& 0.954
& \textbf{0.967}
& 0.924
& \textbf{0.988} \\
GLiNER-multi-PII
& Textract
& 1.000
& 0.000
& 0.950
& 0.973
& \textbf{0.908}
& 0.990 \\
NVIDIA GLiNER PII
& Textract
& 1.000
& 0.000
& \textbf{0.942}
& \textbf{0.967}
& 0.916
& 0.992 \\

\rowcolor{groupgray}
\multicolumn{8}{l}{
\textbf{Full schema coverage without a separate OCR stage}
\hspace{0.4em}\textit{(OCR-free VLMs)}
} \\
Qwen3-VL-32B
& --
& 1.000
& 0.000
& 0.990
& 0.982
& 0.975
& 1.000 \\
InternVL3-38B
& --
& 1.000
& 0.000
& 1.000
& 1.000
& 0.998
& 1.000 \\
GPT-5.4
& --
& 1.000
& 0.000
& 0.998
& 0.988
& 0.990
& 1.000 \\
GPT-5.5
& --
& 1.000
& 0.000
& 0.956
& 0.994
& 0.986
& 1.000 \\
GPT-5.5 + Code Interpreter
& --
& 1.000
& 0.000
& \textbf{0.887}
& \textbf{0.964}
& \textbf{0.908}
& \textbf{0.990} \\

\hline
\end{tabular}

\caption{
Supported-schema leakage at IoU $\tau=0.75$.
Coverage is the fraction of benchmark types supported; Unsup.\ pages
contain at least one unsupported gold type.
Group scores retain only supported gold types.
The full-schema Textract block fixes OCR and schema, isolating the
detector.
Bold marks the best coverage within the restricted-schema block and the
lowest leakage within each full-schema block.
Restricted-schema leakage is not ranked because supported type sets
differ.
Dashes denote no supported group type or no separate OCR stage.
}
\label{tab:supported_schema}
\end{table*}

\subsection{Analysis}
\label{sec:analysis}

\paragraph{Overall safety picture.}
Table~\ref{tab:main_results} shows that stronger systems improve
entity-level localization without achieving document-level safety.
Amazon Comprehend PII obtains the highest localization F1
(F1$_{\mathrm{loc}}=0.304$), while GPT-5.5 with Code Interpreter
obtains the highest typed F1 (F1$_{\mathrm{type}}=0.119$).
Nevertheless, both have DocLeak$_{\mathrm{crit}}=0.968$, showing
that better average detection does not guarantee that a page is safe
to release.

This conclusion is robust to the matching scope. Crediting any
IoU-eligible box regardless of its predicted type reduces the
DocLeak$_{\mathrm{crit}}$ range from 0.968--1.000 under headline
type-aware matching to 0.945--1.000 under type-agnostic matching.
Thus, the high leakage rates are not primarily caused by type-label
errors.

\paragraph{Leakage is severe, not marginal.}
To complement binary DocLeak, we count unmatched critical identifiers per
page. Across systems, leaking pages retain $5.3$--$8.7$ critical identifiers
on average. Even the lowest-severity system misses $5.3$ per leaking page
(median $4$; P90 $11$), corresponding to approximately $66\%$ of critical
identifiers across applicable pages. Thus, leakage typically reflects multiple
exposed identifiers rather than an isolated miss. Complete per-system severity statistics are reported in Table~\ref{tab:severity_diagnostics}.

\paragraph{Schema coverage is not enough.}
Schema coverage diagnostics are reported in Appendix~\ref{app:schema_coverage}. 
Restricted systems such as Presidio, Presidio-best (GLiNER restricted), and Amazon Comprehend PII cover only 20.8\% of benchmark types, while Google DLP and the OpenAI Privacy Filter cover 25.0\%. 
These restricted systems cover Direct identifiers but miss most Linkage keys under our schema mapping, making lookup-style identifiers a persistent coverage gap. 
However, full label coverage does not solve the task either: GLiNER-base, GLiNER multi-PII, and NVIDIA GLiNER PII are configured with full schema coverage, yet their Linkage leakage remains near-total in Appendix~\ref{app:group_leakage}. 
This suggests that sparse lookup keys, dense tables, OCR fragmentation, and layout variation remain difficult even when the label set is available.

\paragraph{Supported-schema leakage isolates detection failures.}
Table~\ref{tab:supported_schema} filters gold labels to the types each system can emit and reports leakage by privacy group.
Even under this easier setting, leakage remains high: full-schema OCR-dependent systems still leak on more than 94\% of Direct pages and more than 96\% of Linkage pages.
This shows that failures are not only taxonomy mismatches; OCR errors, repeated identifiers, span fragmentation, and box reconstruction also contribute to residual exposure.

\paragraph{OCR-dependent failure attribution.}
Full results appear in Appendix Tables~\ref{tab:ocr_attribution}--\ref{tab:ocr_recoverability}. Using type-agnostic matching, we attribute each leaked critical identifier to an OCR miss, a detector miss with zero box overlap, or a box/alignment failure with positive overlap below $\tau=0.75$; one-to-one assignment conflicts account for at most $0.2\%$. Box/alignment failure is the largest category for seven of nine OCR-dependent systems, accounting for $45.9$--$84.5\%$ of their attributable leaks, while detector misses vary from $15.5$ to $48.8\%$. Independently of any detector, Textract cannot recover 623/4,280 value-bearing critical annotations ($14.6\%$), compared with 765/4,280 ($17.9\%$) for Tesseract. Performance differences among the Textract-based pipelines therefore arise primarily downstream of their shared OCR stage. 

\begin{table*}[t]
\centering
\small
\setlength{\tabcolsep}{5pt}
\begin{tabular}{lrrrrrr}
\toprule
\textbf{System}
& \textbf{Valid $\uparrow$}
& \textbf{Leaked crit.}
& \textbf{Interface $\downarrow$}
& \textbf{Zero-IoU $\downarrow$}
& \textbf{Box fail $\downarrow$}
& \textbf{Assignment $\downarrow$} \\
\midrule
Qwen3-VL-32B
& .666 & 4,013 & 27.8 & 20.5 & 51.7 & 0.0 \\
InternVL3-38B
& .540 & 4,274 & 36.9 & 35.1 & 28.0 & 0.0 \\
GPT-5.4 & 1.000 & 4,060 & 0.0 & 16.5 & 83.5 & 0.0 \\
GPT-5.5
& 1.000 & 3,787 & 0.0 & 3.9 & 96.1 & 0.0 \\
GPT-5.5 + Code Interpreter
& 1.000 & 2,549 & 0.0 & 0.9 & 98.9 & 0.2 \\
\bottomrule
\end{tabular}
\caption{
Error attribution for leaked Direct and Linkage identifiers from OCR-free
VLMs at $\operatorname{IoU}=0.75$, using type-agnostic matching.
Attribution columns are percentages of the leaked critical instances shown
in the second numeric column and sum to 100\%.
Interface denotes identifiers on pages without a fully valid output;
Zero-IoU denotes no overlapping prediction; Box fail denotes positive
overlap below the IoU threshold; and Assignment denotes one-to-one
matching conflicts.
}
\label{tab:ocr-free-attribution}
\end{table*}

\paragraph{Formatting versus localization in OCR-free VLMs.}
Table~\ref{tab:ocr-free-attribution} separates interface failures from
failures on valid outputs using a common denominator of leaked critical
identifiers. Interface failures account for 27.8\% and 36.9\% of
leakage for Qwen3-VL-32B and InternVL3-38B, respectively, but none for
GPT-5.4 or either GPT-5.5 configuration. Among residual leaked critical
instances, sub-threshold boxes account for 51.7\% for Qwen3-VL-32B,
83.5\% for GPT-5.4, 96.1\% for GPT-5.5, and 98.9\% for GPT-5.5 with
Code Interpreter. These percentages describe the composition of the
errors that remain, not the overall localization-failure rate. A larger
box-failure share can therefore coexist with improved
F1$_{\mathrm{loc}}$ when total leakage decreases and zero-overlap
misses decline more sharply. InternVL3-38B instead exhibits a larger
share of zero-overlap misses (35.1\%). Thus, interface reliability is
necessary but insufficient; spatial localization remains the dominant
limitation for the GPT configurations and Qwen3-VL-32B.

\begin{figure}[!t]
\centering
\includegraphics[width=\columnwidth]{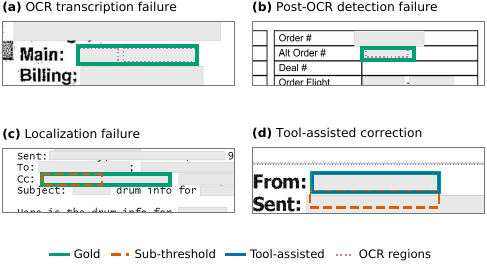}
\caption{Representative failure modes. Panels (a)--(c) use
GLiNER-multi-PII with Textract; panel (d) compares GPT-5.5
single-pass and tool-assisted localization. Green denotes gold
boxes, orange dashed boxes sub-threshold predictions, blue
tool-assisted predictions, and magenta dotted boxes OCR regions.}
\label{fig:qualitative_errors}
\end{figure}

\paragraph{Tool assistance improves OCR-free localization, but leakage remains high.}
Figure~\ref{fig:qualitative_errors} illustrates these failures and a
tool-assisted localization correction. Within GPT-5.5, Code
Interpreter raises F1$_{\mathrm{loc}}$ from 0.090 to 0.249 and
F1$_{\mathrm{type}}$ from 0.050 to 0.119. It reduces
DocLeak$_{\mathrm{crit}}$ from 0.990 to 0.968 and reduces Direct and
Contextual leakage from 0.956 to 0.887 and from 0.986 to 0.908,
respectively. Linkage leakage nevertheless remains 0.964.
Tool-assisted inspection therefore improves spatial grounding
substantially, but does not yet provide release-level redaction safety.

\paragraph{Takeaway.}
LeakageBench is best viewed as a redaction-safety challenge set rather
than a leaderboard where systems differ only by small margins. The
benchmark exposes complementary failure modes: restricted taxonomies
omit identifier groups, OCR-dependent pipelines propagate recognition
and alignment errors, and OCR-free VLMs remain limited by spatial
grounding. Code Interpreter substantially improves GPT-5.5
localization, but the resulting system still leaks critical PII on
96.8\% of applicable pages. These findings motivate systems that
combine broad schema coverage, robust visual reading, and tight,
high-recall localization.

\section*{Ethical, Legal, and Release Protocol}

LeakageBench is intended for defensive research on document-redaction
safety. Although the source documents originate from public
collections, combining real document images with fine-grained PII
annotations introduces privacy and dual-use risks. The document images
and annotations will therefore be made available for research use under
a privacy- and license-aware Data Use Agreement (DUA).

The benchmark is intended for evaluating privacy-preserving redaction,
not for re-identification or PII mining. Its GDPR-aligned schema
supports comparative evaluation but does not define legal compliance
or prescribe which fields must be redacted in every jurisdiction or
deployment setting.

\section*{Limitations}

LeakageBench is a 500-page challenge set drawn from public business
documents, not a representative sample of all redaction settings. It
does not cover heavily handwritten, medical, legal-discovery,
multilingual, or non-U.S. documents. Evaluation is page-local and
does not measure multi-page contextual inference.

The schema follows an identifiability-first view. Direct identifiers
and Linkage keys are primary release-safety targets, whereas masking
Contextual identifiers depends on the applicable policy.

Headline DocLeak uses conservative type-aware matching at IoU
$\tau=0.75$, so a wrongly typed but correctly localized box counts as
a leak. However, type-agnostic matching yields the same substantive
conclusion, with DocLeak$_{\mathrm{crit}}$ remaining between 0.945
and 1.000. Axis-aligned boxes and IoU do not directly measure pixel
coverage of skewed, handwritten, or irregular regions. Although F1
penalizes extra predictions, we do not define a document-level
over-redaction or utility metric.

Error attribution is limited by the stored artifacts: 35 critical
annotations lack usable verbatim values, span-level alignment logs
are unavailable, and Google DLP's vendor OCR is not stored.
Consequently, box and alignment errors cannot be separated fully.
The OCR-free results also represent specific model and inference
configurations whose API behavior may change.

\bibliography{custom}

\appendix

\section{Annotation Instructions}
\label{app:annotation_instructions}

These instructions describe how to annotate \textbf{LeakageBench} for privacy redaction. The goal is to mark \emph{every visible region} on a page that contains an in-scope identifier, using tight bounding boxes and an exact transcription of the text.

\subsection{What to label (schema)}
Annotate \textbf{all} instances of the following fine-grained PII types (must match exactly):

\begin{itemize}
    \item \textbf{Direct identifiers:} \textsc{Person Name, Address, Email Address, Phone Number}
    \item \textbf{Linkage keys:} \textsc{Customer ID, Employee ID, Policy Number, ID Card Number, File Number, Reference Number, Contract Number, Invoice Number, Cheque Number, Registration Number, Call Sign, DA Case Number}
    \item \textbf{Operational references:} \textsc{Agency Code, Agency Ref, Advertiser Ref}
    \item \textbf{Contextual identifiers:} \textsc{Date, Organization Name, Business ID, Location}
    \item \textbf{Other (visual PII):} \textsc{Signature} (handwritten signatures or signature marks)
\end{itemize}

\paragraph{Important:} If a page contains multiple identifiers, annotate \emph{all of them}. This benchmark is safety-driven: a page is considered leaked if \emph{any} in-scope identifier is missed.

\subsection{How to draw boxes (localization)}
\begin{itemize}
    \item Draw a \textbf{tight bounding box} around the \textbf{visible identifier text} (or signature mark). Boxes should cover the characters with minimal background.
    \item Do \textbf{not} box entire rows, table blocks, or paragraphs unless \emph{the entire region is PII}.
    \item If an identifier spans multiple lines (e.g., an address block), draw \textbf{one box covering the full multi-line span} if the tool supports it. Otherwise, use multiple boxes and keep the transcription consistent across lines.
    \item If the identifier is repeated on the page (e.g., header and footer), annotate \textbf{each occurrence} separately.
\end{itemize}

\subsection{How to enter the text value (transcription)}
\begin{itemize}
    \item Transcribe the \textbf{value exactly as it appears} in the image.
    \item Do \textbf{not} correct spelling, punctuation, casing, or spacing.
    \item Do \textbf{not} normalize formats (dates, phone numbers, IDs, etc.). Keep the original formatting.
    \item For multi-line values, use the literal token \texttt{\textbackslash n} to indicate line breaks (e.g., \texttt{Line1\textbackslash nLine2\textbackslash nLine3}).
\end{itemize}

\subsection{Type selection rules (common edge cases)}
Use the most specific schema type that matches the \emph{role} of the text on the page.

\begin{itemize}
    \item \textbf{Person Name:} Names of individuals (signers, recipients, customers, employees). If the page contains initials, titles, or suffixes as part of the displayed name, include them as shown.
    \item \textbf{Organization Name:} Company, agency, or institution names. Do not label generic department words unless they clearly function as an entity name in context.
    \item \textbf{Address:} Postal addresses (street + city/state/zip) or address blocks. Include apartment/suite/unit numbers if present.
    \item \textbf{Email Address:} Email addresses exactly as written.
    \item \textbf{Phone Number:} Telephone numbers (including extensions if shown).
    \item \textbf{Date:} Dates and date ranges as written (including billing periods and coverage periods).
    \item \textbf{Invoice Number / Contract Number / Policy Number / Reference Number / File Number:} Choose the type based on the field label and context. If the page label says “Invoice \#”, annotate as \textsc{Invoice Number}. If it says “Policy No.”, annotate as \textsc{Policy Number}, etc.
    \item \textbf{Customer ID / Employee ID / Business ID:} Use when the page explicitly indicates the identifier category (e.g., “Customer ID”, “Employee ID”, “UEN/ABN/VAT”, “Business ID”).
    \item \textbf{Call Sign:} Broadcast or station identifiers (e.g., \texttt{KXAS}, \texttt{WLAX/WEUX}) when used as a station/call sign.
    \item \textbf{Agency Code / Agency Ref / Advertiser Ref / DA Case Number:} Use when the page contains explicit labeled codes corresponding to these identifiers.
    \item \textbf{Signature:} Handwritten signature marks or signature scribbles. Do not label printed names near the signature line as \textsc{Signature};
\end{itemize}

\paragraph{When unsure.}
If you cannot confidently map a span to one of the schema types, \textbf{do not guess}. Prefer leaving it unannotated over inventing a type.

\subsection{Pre-marked / highlighted regions}
Some pages may contain regions already highlighted or redacted (e.g., dark blocks, overlays, pre-annotations). These are \textbf{not reliable ground-truth evidence of PII}.
\begin{itemize}
    \item Do \textbf{not} annotate pre-redacted regions where the text is no longer readable.
    \item If the text remains visible despite highlighting, annotate it normally based on what is visible.
\end{itemize}

\subsection{Quality checklist before submitting a page}
Before finalizing a page, quickly verify:
\begin{itemize}
    \item All visible names, emails, phone numbers, and addresses are boxed.
    \item All visible linkage keys (invoice/contract/policy/reference/file/customer IDs, etc.) are boxed.
    \item Boxes are tight and not overly large.
    \item Transcriptions are verbatim and include \texttt{\textbackslash n} for multi-line items.
\end{itemize}

\section{Baseline Reproducibility Details}
\label{app:baseline_details}

This appendix records the implementation choices needed to reproduce the baseline evaluations. 
No baseline is trained or tuned on LeakageBench. 
All outputs are converted to the same format: PII type, optional value, confidence score, and image-space bounding box.

\begin{table*}[t]
\centering
\small
\setlength{\tabcolsep}{4pt}
\renewcommand{\arraystretch}{1.08}
\begin{tabular}{l l l l}
\hline
\textbf{System} & \textbf{OCR / boxes} & \textbf{Detector} & \textbf{Type scope} \\
\hline
Presidio & Tesseract & spaCy \texttt{en\_core\_web\_lg} & mapped built-in types \\
Presidio-best & Textract & GLiNER restricted & restricted PII types \\
GLiNER-base & Textract & GLiNER-base & full LeakageBench schema \\
GLiNER-multi-PII & Textract & GLiNER multi-PII & PII-specialized types \\
NVIDIA GLiNER PII & Textract & NVIDIA GLiNER PII & PII-specialized types \\
GLiNER2 & Textract & GLiNER2 & mapped schema (23/24 types) \\
Google DLP & service/OCR text & Google DLP & mapped built-in types \\
Amazon Comprehend PII & Textract & Amazon Comprehend PII & mapped built-in types \\
OpenAI Privacy Filter & Textract & OpenAI Privacy Filter & mapped built-in types \\
Qwen3-VL-32B & image & Qwen3-VL-32B & prompted full schema \\
InternVL3-38B & image & InternVL3-38B & prompted full schema \\
GPT-5.4 & image & GPT-5.4 & prompted full schema \\
GPT-5.5 & image & GPT-5.5 & prompted full schema \\
GPT-5.5 + Code Interpreter & image + tool inspection & GPT-5.5 & prompted full schema \\
\hline
\end{tabular}
\caption{Baseline inventory. ``OCR / boxes'' indicates the source of text and localization boxes. ``Type scope'' indicates whether the system uses built-in labels mapped to LeakageBench, a restricted PII label set, or the full benchmark schema.}
\label{tab:baseline_inventory}
\end{table*}

Type scope controls which LeakageBench labels a system can emit before scoring; it is used only for schema coverage and supported-type leakage, not for full-schema evaluation.

\paragraph{OCR-dependent conversion.}
OCR-dependent systems detect PII over recognized text and recover page coordinates from OCR word boxes. 
Presidio uses Tesseract OCR. 
The stronger OCR-dependent baselines use Amazon Textract word boxes, except for services that provide their own text recognition or span interface. 
For each detected text span, we align the span to OCR words and use the minimal enclosing rectangle over the matched word boxes. 
Unaligned detections are discarded.

\paragraph{Schema mapping and unsupported types.}
All predictions are mapped to the LeakageBench schema before scoring. 
For systems with built-in or restricted taxonomies, unmapped prediction labels are discarded. 
Under full-schema evaluation, unsupported gold types remain in scope and count as missed detections. 
Supported-type leakage filters gold labels to the entity types each system can emit.

\paragraph{OCR-free conversion and validation.}
OCR-free VLMs receive the page image directly and output normalized boxes, which we convert to pixel coordinates before scoring. 
The full LeakageBench schema is supplied in the prompt, so OCR-free schema coverage is $1.0$ by configuration. 
Malformed outputs, labels outside the schema, invalid boxes, and out-of-bounds boxes are scored as empty predictions; we do not apply model-specific repair or manual correction.

\section{Prompt Templates}
\label{app:prompts}

\subsection{Shared prompt header}
\label{app:prompt-header}
\begin{lstlisting}[style=promptstyle,caption={Shared prompt header used for all models.},label={lst:prompt_header}]
PROMPT_HEADER = r"""
You are a privacy redaction detector for BUSINESS DOCUMENT IMAGES (emails, invoices, forms, letters).

TASK
Detect and localize ALL regions that contain PII (Personally Identifiable Information) on this single page.

DRAWING BOXES
Draw boxes tightly around the visible PII text (do not box entire rows/tables unless the whole region is PII).

SCORE
Set score to your confidence [0,1] that (a) the box tightly encloses the PII text and (b) the type is correct; use lower scores for uncertain cases, and if you include a prediction you must still provide a score

ALLOWED PII TYPES (MUST MATCH EXACTLY)
{schema_types}

TYPE CONSTRAINT (IMPORTANT)
- You MUST output ONLY the allowed PII types listed above.
- Do NOT output any other labels (no synonyms, no extra categories).
- If you are unsure of the correct allowed type, OMIT the prediction.

NO-PII CASE
If the page contains no PII of the allowed types, follow the NO-PII output rule in OUTPUT FORMAT.

RULES
- Output ONLY what is required by OUTPUT FORMAT (no prose, no Markdown).
- Do NOT invent text for "value". If unsure, use "".
"""
\end{lstlisting}

\subsection{Qwen prompt template (JSON output)}
\label{app:prompt-qwen}
\begin{lstlisting}[style=promptstyle,caption={Prompt template for Qwen-style models with strict JSON output.},label={lst:prompt_qwen}]
PROMPT_TEMPLATE_QWEN = PROMPT_HEADER + r"""
OUTPUT FORMAT (STRICT)
Return exactly one JSON object:
{
  "predictions": [
    {"bbox_2d":[x1,y1,x2,y2], "type":"<one allowed type>", "score":0.0, "value":""}
  ]
}

COORDINATES
- bbox_2d MUST be normalized to [0,1000] with origin at top-left.
- Each box is [x1,y1,x2,y2] with x1<x2 and y1<y2.

NO-PII OUTPUT
{"predictions": []}
"""
\end{lstlisting}

\subsection{InternVL prompt template (tagged blocks)}
\label{app:prompt-internvl}
\begin{lstlisting}[style=promptstyle,caption={Prompt template for InternVL-style models with tagged block outputs.},label={lst:prompt_internvl}]
PROMPT_TEMPLATE_INTERNVL = PROMPT_HEADER + r"""
EXAMPLE (one block):
<ref>Date</ref><box>[[120,340,210,360]]</box><value>01/28/20</value><score>0.95</score>

OUTPUT FORMAT (STRICT)
Return ONLY plain text consisting of zero or more repeated blocks:
<ref>TYPE</ref><box>[[x1,y1,x2,y2], ...]</box><value>...</value><score>...</score>

COORDINATES
- Boxes MUST be normalized to [0,1000] with origin at top-left.
- Each box is [x1,y1,x2,y2] with x1<x2 and y1<y2.

NO-PII OUTPUT
Output an EMPTY string (no characters).
"""
\end{lstlisting}

\section{Document-Type Composition}
\label{app:doc_types}
Table ~\ref{tab:doc_types_app} reports the page-type distribution for each source. The benchmark is dominated by invoices, emails, forms, correspondence, and file-like pages, reflecting the operational-document focus of LeakageBench.

\begin{table*}[t]
\centering
\small
\setlength{\tabcolsep}{5pt}
\renewcommand{\arraystretch}{1.08}
\begin{tabular}{l r r r r r r}
\hline
\textbf{Source} & \textbf{Invoice} & \textbf{Forms} & \textbf{Email(s)} & \textbf{Letter/Corr.} & \textbf{File} & \textbf{Other} \\
\hline
Source A (OCR-IDL) & 3 & 2 & 22 & 46 & 30 & 10 \\
Source B (VRDU Ad-Buy) & 219 & 2 & 0 & 0 & 1 & 0 \\
Source C (FCC free-form) & 7 & 35 & 68 & 16 & 16 & 23 \\
Total & 229 & 39 & 90 & 62 & 47 & 33 \\
\hline
\end{tabular}
\caption{Page-type composition by source. ``Email(s)'' aggregates Email and Emails; ``Letter/Corr.'' aggregates Letter and Correspondence.}
\label{tab:doc_types_app}
\end{table*}

\section{Schema Coverage}
\label{app:schema_coverage}

Table~\ref{tab:schema_coverage_appendix} reports entity-type coverage
under each baseline's schema mapping or prompting interface. Coverage
indicates which LeakageBench types the evaluated configuration was
allowed to emit; it does not measure intrinsic model capability or
detection accuracy.

\begin{table*}[t]
\centering
\small
\setlength{\tabcolsep}{6pt}
\renewcommand{\arraystretch}{1.12}
\begin{tabular}{l r r r r}
\hline
\textbf{System} &
\textbf{All Types$\uparrow$} &
\textbf{Direct$\uparrow$} &
\textbf{Linkage$\uparrow$} &
\textbf{Contextual$\uparrow$} \\
\hline

\multicolumn{5}{l}{\textit{OCR-dependent pipelines}} \\
Presidio & 0.208 & 1.000 & 0.000 & 0.333 \\
Presidio-best & 0.208 & 1.000 & 0.000 & 0.333 \\
GLiNER-base & 1.000 & 1.000 & 1.000 & 1.000 \\
GLiNER-multi-PII & 1.000 & 1.000 & 1.000 & 1.000 \\
NVIDIA GLiNER PII & 1.000 & 1.000 & 1.000 & 1.000 \\
GLiNER2 & 0.958 & 1.000 & 1.000 & 1.000 \\
Google DLP & 0.250 & 1.000 & 0.000 & 0.667 \\
Amazon Comprehend PII & 0.208 & 1.000 & 0.000 & 0.333 \\
OpenAI Privacy Filter & 0.250 & 1.000 & 0.077 & 0.333 \\

\hline
\multicolumn{5}{l}{\textit{OCR-free VLMs}} \\
Qwen3-VL-32B & 1.000 & 1.000 & 1.000 & 1.000 \\
InternVL3-38B & 1.000 & 1.000 & 1.000 & 1.000 \\
GPT-5.4                    & 1.000 & 1.000 & 1.000 & 1.000 \\
GPT-5.5                    & 1.000 & 1.000 & 1.000 & 1.000 \\
GPT-5.5 + Code Interpreter & 1.000 & 1.000 & 1.000 & 1.000 \\

\hline
\end{tabular}
\caption{Schema coverage of each evaluated configuration after fixed
schema mapping or prompting. Coverage reflects the prediction interface,
not intrinsic model capability or detection accuracy.}
\label{tab:schema_coverage_appendix}
\end{table*}

\section{Group-Level Leakage}
\label{app:group_leakage}

Table~\ref{tab:group_leakage_app} reports conditional DocLeak for each privacy-aligned group, computed over pages containing at least one ground-truth identifier from that group.

\begin{table*}[t]
\centering
\small
\setlength{\tabcolsep}{6pt}
\renewcommand{\arraystretch}{1.12}
\begin{tabular}{l r r r}
\hline
\textbf{System} &
\textbf{Direct$\downarrow$} &
\textbf{Linkage$\downarrow$} &
\textbf{Contextual$\downarrow$} \\
\hline

\multicolumn{4}{l}{\textit{OCR-dependent pipelines}} \\
Presidio & 0.973 & 1.000 & 0.982 \\
Presidio-best & 0.967 & 1.000 & 0.977 \\
GLiNER-base & 0.954 & 0.967 & 0.924 \\
GLiNER-multi-PII & 0.950 & 0.973 & 0.908 \\
NVIDIA GLiNER PII & 0.942 & 0.967 & 0.916 \\
GLiNER2 & 0.971 & 0.985 & 0.963 \\
Google DLP & 1.000 & 1.000 & 0.990 \\
Amazon Comprehend PII & 0.908 & 1.000 & 0.971 \\
OpenAI Privacy Filter & 0.971 & 1.000 & 0.994 \\

\hline
\multicolumn{4}{l}{\textit{OCR-free vision-language models}} \\
Qwen3-VL-32B & 0.990 & 0.982 & 0.975 \\
InternVL3-38B & 1.000 & 1.000 & 0.998 \\
GPT-5.4                    & 0.998 & 0.988 & 0.990 \\
GPT-5.5                    & 0.956 & 0.994 & 0.986 \\
GPT-5.5 + Code Interpreter & 0.887 & 0.964 & 0.908 \\

\hline
\end{tabular}

\caption{Conditional document leakage by privacy group at IoU $\tau=0.75$. 
Each column reports DocLeak over pages containing at least one ground-truth identifier from that group. 
All systems use full-schema scoring; unsupported types and invalid outputs count as misses.}
\label{tab:group_leakage_app}
\end{table*}

\section{IoU Threshold Sweep}
\label{app:iou_sweep}

Table~\ref{tab:iou_sweep} reports a threshold sweep over the IoU matching criterion $\tau \in \{0.50,0.75,0.95\}$. 
Lowering $\tau$ relaxes box tightness requirements, while higher $\tau$ requires tighter spatial alignment. 
Although entity-level F1 changes substantially with $\tau$, document-level leakage remains high across thresholds.

\begin{table*}[t]
\centering
\small
\setlength{\tabcolsep}{4.5pt}
\renewcommand{\arraystretch}{1.06}
\begin{tabular}{l r r r r r}
\toprule
\textbf{System} & \textbf{IoU} &
\textbf{F1$_{\text{loc}}$} &
\textbf{F1$_{\text{type}}$} &
\textbf{DocLeak$_{\text{all}}\downarrow$} &
\textbf{DocLeak$_{\text{crit}}\downarrow$} \\
\midrule

\multirow{3}{*}{Presidio}
& 0.50 & 0.448 & 0.109 & 0.992 & 0.964 \\
& \cellcolor{gray!10}\textbf{0.75}
& \cellcolor{gray!10}0.137
& \cellcolor{gray!10}0.048
& \cellcolor{gray!10}0.994
& \cellcolor{gray!10}0.990 \\
& 0.95 & 0.001 & 0.001 & 1.000 & 1.000 \\
\midrule

\multirow{3}{*}{Presidio-best}
& 0.50 & 0.529 & 0.110 & 0.992 & 0.968 \\
& \cellcolor{gray!10}\textbf{0.75}
& \cellcolor{gray!10}0.248
& \cellcolor{gray!10}0.064
& \cellcolor{gray!10}0.996
& \cellcolor{gray!10}0.986 \\
& 0.95 & 0.020 & 0.004 & 1.000 & 1.000 \\
\midrule

\multirow{3}{*}{GLiNER-base}
& 0.50 & 0.403 & 0.136 & 0.988 & 0.949 \\
& \cellcolor{gray!10}\textbf{0.75}
& \cellcolor{gray!10}0.216
& \cellcolor{gray!10}0.063
& \cellcolor{gray!10}0.988
& \cellcolor{gray!10}0.978 \\
& 0.95 & 0.031 & 0.004 & 1.000 & 1.000 \\
\midrule

\multirow{3}{*}{GLiNER-multi-PII}
& 0.50 & 0.432 & 0.184 & 0.974 & 0.939 \\
& \cellcolor{gray!10}\textbf{0.75}
& \cellcolor{gray!10}0.245
& \cellcolor{gray!10}0.091
& \cellcolor{gray!10}0.990
& \cellcolor{gray!10}0.972 \\
& 0.95 & 0.030 & 0.005 & 1.000 & 1.000 \\
\midrule

\multirow{3}{*}{NVIDIA GLiNER PII}
& 0.50 & 0.567 & 0.251 & 0.992 & 0.972 \\
& \cellcolor{gray!10}\textbf{0.75}
& \cellcolor{gray!10}0.288
& \cellcolor{gray!10}0.109
& \cellcolor{gray!10}0.992
& \cellcolor{gray!10}0.986 \\
& 0.95 & 0.026 & 0.004 & 1.000 & 1.000 \\
\midrule

\multirow{3}{*}{GLiNER2}
& 0.50 & 0.311 & 0.139 & 0.994 & 0.976 \\
& \cellcolor{gray!10}\textbf{0.75}
& \cellcolor{gray!10}0.156
& \cellcolor{gray!10}0.065
& \cellcolor{gray!10}1.000
& \cellcolor{gray!10}0.994 \\
& 0.95 & 0.018 & 0.004 & 1.000 & 1.000 \\
\midrule

\multirow{3}{*}{Google DLP}
& 0.50 & 0.286 & 0.077 & 0.994 & 0.976 \\
& \cellcolor{gray!10}\textbf{0.75}
& \cellcolor{gray!10}0.076
& \cellcolor{gray!10}0.013
& \cellcolor{gray!10}1.000
& \cellcolor{gray!10}1.000 \\
& 0.95 & 0.000 & 0.000 & 1.000 & 1.000 \\
\midrule

\multirow{3}{*}{Amazon Comprehend PII}
& 0.50 & 0.589 & 0.136 & 0.976 & 0.901 \\
& \cellcolor{gray!10}\textbf{0.75}
& \cellcolor{gray!10}0.304
& \cellcolor{gray!10}0.085
& \cellcolor{gray!10}0.992
& \cellcolor{gray!10}0.968 \\
& 0.95 & 0.033 & 0.006 & 1.000 & 0.998 \\
\midrule

\multirow{3}{*}{OpenAI Privacy Filter}
& 0.50 & 0.272 & 0.086 & 0.990 & 0.964 \\
& \cellcolor{gray!10}\textbf{0.75}
& \cellcolor{gray!10}0.164
& \cellcolor{gray!10}0.055
& \cellcolor{gray!10}0.996
& \cellcolor{gray!10}0.990 \\
& 0.95 & 0.024 & 0.005 & 1.000 & 1.000 \\
\midrule

\multirow{3}{*}{Qwen3-VL-32B}
& 0.50 & 0.224 & 0.134 & 0.992 & 0.978 \\
& \cellcolor{gray!10}\textbf{0.75}
& \cellcolor{gray!10}0.073
& \cellcolor{gray!10}0.046
& \cellcolor{gray!10}1.000
& \cellcolor{gray!10}0.998 \\
& 0.95 & 0.001 & 0.000 & 1.000 & 1.000 \\
\midrule

\multirow{3}{*}{InternVL3-38B}
& 0.50 & 0.090 & 0.048 & 1.000 & 0.996 \\
& \cellcolor{gray!10}\textbf{0.75}
& \cellcolor{gray!10}0.012
& \cellcolor{gray!10}0.019
& \cellcolor{gray!10}1.000
& \cellcolor{gray!10}1.000 \\
& 0.95 & 0.000 & 0.000 & 1.000 & 1.000 \\
\midrule

\multirow{3}{*}{GPT-5.4}
& 0.50 & 0.242 & 0.183 & 0.988 & 0.953 \\
& \cellcolor{gray!10}\textbf{0.75}
& \cellcolor{gray!10}0.044
& \cellcolor{gray!10}0.031
& \cellcolor{gray!10}1.000
& \cellcolor{gray!10}1.000 \\
& 0.95 & 0.000 & 0.000 & 1.000 & 1.000 \\
\midrule

\multirow{3}{*}{GPT-5.5}
& 0.50 & 0.369 & 0.248 & 0.958 & 0.907 \\
& \cellcolor{gray!10}\textbf{0.75}
& \cellcolor{gray!10}0.090
& \cellcolor{gray!10}0.050
& \cellcolor{gray!10}1.000
& \cellcolor{gray!10}0.990 \\
& 0.95 & 0.001 & 0.000 & 1.000 & 1.000 \\
\midrule

\multirow{3}{*}{GPT-5.5 + Code Interpreter}
& 0.50 & 0.614 & 0.380 & 0.904 & 0.775 \\
& \cellcolor{gray!10}\textbf{0.75}
& \cellcolor{gray!10}0.249
& \cellcolor{gray!10}0.119
& \cellcolor{gray!10}0.990
& \cellcolor{gray!10}0.968 \\
& 0.95 & 0.009 & 0.003 & 1.000 & 1.000 \\

\bottomrule
\end{tabular}
\caption{IoU threshold sweep under full-schema scoring. Lower IoU thresholds relax box alignment, while higher thresholds require tighter localization. Shaded metric cells correspond to the main-paper threshold $\tau=0.75$.}
\label{tab:iou_sweep}
\end{table*}

\section{Corpus and Entity Statistics}
\label{app:entity-stats}

Table~\ref{tab:app-corpus-entity-stats} reports corpus-level entity-category statistics by source, and Table~\ref{tab:app-entity-type-stats} reports fine-grained entity-type counts.

\section{Additional Error Diagnostics}
\label{app:error_diagnostics}

Table~\ref{tab:severity_diagnostics} reports per-system leakage
severity using the headline type-aware matcher.
Table~\ref{tab:ocr_attribution} attributes OCR-dependent leakage
using type-agnostic matching, while
Table~\ref{tab:ocr_recoverability} measures detector-independent
OCR recoverability using only gold values and stored OCR text.

\begin{table*}[t]
\centering
\small
\setlength{\tabcolsep}{6pt}
\renewcommand{\arraystretch}{1.15}
\begin{tabular}{lrrrrrrrrr}
\toprule
& \multicolumn{2}{c}{\textbf{Corpus}} & \multicolumn{2}{c}{\textbf{Mentions}} & \multicolumn{5}{c}{\textbf{By Category}} \\
\cmidrule(lr){2-3}\cmidrule(lr){4-5}\cmidrule(lr){6-10}
\textbf{Src} & \textbf{Docs} & \textbf{Pgs} & \textbf{All} & \textbf{/Pg} & \textbf{DI} & \textbf{CI} & \textbf{LK} & \textbf{Sig} & \textbf{OR} \\
\midrule
A   & 103 & 113 & 1714  & 15.2 & 1143 &  445 & 109 & 17 &   0 \\
B   & 107 & 222 & 7331  & 33.0 &  673 & 6245 & 221 &  5 & 187 \\
C   &  81 & 165 & 2909  & 17.6 & 1963 &  710 & 206 & 30 &   0 \\
\midrule
All & 291 & 500 & 11954 & 23.9 & 3779 & 7400 & 536 & 52 & 187 \\
\bottomrule
\end{tabular}

\caption{Corpus and entity-category statistics by source. DI = Direct Identifiers, CI = Contextual Identifiers, LK = Linkage Keys, Sig = Signature, OR = Operational References (Agency Code + Agency Reference + Advertiser Reference).}
\label{tab:app-corpus-entity-stats}
\end{table*}


\begin{table*}[t]
\centering
\small
\setlength{\tabcolsep}{6pt}
\renewcommand{\arraystretch}{1.05}
\begin{tabular}{llrrrr}
\toprule
\textbf{Group} & \textbf{Entity Type} & \textbf{Src A} & \textbf{Src B} & \textbf{Src C} & \textbf{All} \\
\midrule

\multicolumn{2}{l}{\textbf{Direct Identifiers} (Subtotal)} & 1,143 & 673 & 1,963 & 3,779 \\
\quad & Person Name   & 804 & 238 & 895 & 1,937 \\
\quad & Address       & 156 & 226 & 318 & 700 \\
\quad & Phone Number  & 117 & 205 & 165 & 487 \\
\quad & Email Address &  66 &   4 & 585 & 655 \\

\addlinespace
\multicolumn{2}{l}{\textbf{Contextual Identifiers} (Subtotal)} & 445 & 6,245 & 710 & 7,400 \\
\quad & Date              & 207 & 5,784 & 499 & 6,490 \\
\quad & Organization Name & 238 &   461 & 207 & 906 \\
\quad & Business ID       &   0 &     0 &   4 & 4 \\

\addlinespace
\multicolumn{2}{l}{\textbf{Linkage Keys} (Subtotal)} & 109 & 221 & 206 & 536 \\
\quad & Linkage ID           &  20 & 110 & 109 & 239 \\
\quad & Contract Number      &  68 &  23 &   7 & 98 \\
\quad & Invoice Number       &  13 &  88 &   7 & 108 \\
\quad & ID Card Number       &   8 &   0 &   0 & 8 \\
\quad & Registration Number  &   0 &   0 &  32 & 32 \\
\quad & Reference Number     &   0 &   0 &  19 & 19 \\
\quad & File Number          &   0 &   0 &   9 & 9 \\
\quad & Cheque Number        &   0 &   0 &   8 & 8 \\
\quad & DA Case Number       &   0 &   0 &   5 & 5 \\
\quad & Call Sign            &   0 &   0 &   4 & 4 \\
\quad & Customer ID          &   0 &   0 &   4 & 4 \\
\quad & Employee ID          &   0 &   0 &   1 & 1 \\
\quad & Policy Number        &   0 &   0 &   1 & 1 \\

\addlinespace
\multicolumn{2}{l}{\textbf{Operational References} (Subtotal)} & 0 & 187 & 0 & 187 \\
\quad & Agency Code          & 0 &  45 & 0 & 45 \\
\quad & Agency Reference     & 0 &  75 & 0 & 75 \\
\quad & Advertiser Reference & 0 &  67 & 0 & 67 \\

\addlinespace
\multicolumn{2}{l}{\textbf{Signature} (Subtotal)} & 17 & 5 & 30 & 52 \\
\quad & Signature & 17 & 5 & 30 & 52 \\

\midrule
\textbf{Total} & \textbf{All Entities} & 1,714 & 7,331 & 2,909 & 11,954 \\
\bottomrule
\end{tabular}
\caption{Entity-type statistics by source. Casing is standardized to Title Case; acronyms (e.g., ID, DA) remain uppercase.}
\label{tab:app-entity-type-stats}
\end{table*}


\begin{table*}[t]
\centering
\small
\setlength{\tabcolsep}{7pt}
\renewcommand{\arraystretch}{1.10}

\begin{tabular}{l r r r r r}
\hline
\textbf{System} &
\textbf{DocLeak$_{\mathrm{crit}}\downarrow$} &
\shortstack{\textbf{Mean}\\\textbf{missed$\downarrow$}} &
\textbf{Median$\downarrow$} &
\textbf{P90$\downarrow$} &
\shortstack{\textbf{Mean proportion}\\\textbf{missed$\downarrow$}} \\
\hline

\rowcolor{groupgray}
\multicolumn{6}{l}{\textbf{OCR-dependent pipelines}} \\

Presidio
& 0.990 & 7.002 & 5 & 15 & 0.839 \\

Presidio-best
& 0.986 & 6.639 & 5 & 14 & 0.805 \\

GLiNER-base
& 0.978 & 6.768 & 4 & 14 & 0.785 \\

GLiNER-multi-PII
& 0.972 & 7.013 & 4 & 14 & 0.772 \\

NVIDIA GLiNER PII
& 0.986 & 6.680 & 4 & 14 & 0.767 \\

GLiNER2
& 0.994 & 6.835 & 5 & 15 & 0.827 \\

Google DLP
& 1.000 & 8.425 & 6 & 17.7 & 0.978 \\

Amazon Comprehend PII
& \textbf{0.968}
& \textbf{5.308}
& \textbf{4}
& \textbf{11}
& 0.661 \\

OpenAI Privacy Filter
& 0.990 & 6.196 & 5 & 12 & 0.813 \\

\rowcolor{groupgray}
\multicolumn{6}{l}{\textbf{OCR-free vision-language models}} \\

Qwen3-VL-32B
& 0.998 & 8.183 & 5 & 18 & 0.935 \\

InternVL3-38B
& 1.000 & 8.652 & 6 & 18 & 0.990 \\

GPT-5.4
& 1.000 & 8.245 & 5 & 18 & 0.939 \\

GPT-5.5
& 0.990 & 7.810 & 5 & 16 & 0.860 \\

GPT-5.5 + Code Interpreter
& \textbf{0.968}
& \textbf{5.538}
& \textbf{4}
& \textbf{12}
& \textbf{0.656} \\

\hline
\end{tabular}

\caption{Severity of missed Direct and Linkage identifiers at
IoU $\tau=0.75$. DocLeak and missed counts use the headline
type-aware matcher. Mean, median, and P90 are computed over leaking
pages; mean proportion is computed over all pages containing critical
identifiers. Bold marks the best value within each system family.}
\label{tab:severity_diagnostics}
\end{table*}

\begin{table*}[t]
\centering
\small
\setlength{\tabcolsep}{9pt}
\renewcommand{\arraystretch}{1.12}

\begin{tabular}{l r r r r r}
\hline
\textbf{System} &
\textbf{Attrib. $n$} &
\multicolumn{4}{c}{\textbf{Share of attributable critical leaks (\%)}} \\
\cline{3-6}
&
&
\textbf{OCR miss} &
\textbf{Detector miss} &
\textbf{Box/alignment} &
\textbf{Assignment} \\
\hline

\rowcolor{groupgray}
\multicolumn{6}{l}{
\textbf{Tesseract OCR}
} \\

Presidio
& 3,371
& 21.4
& 21.8
& 56.8
& 0.0 \\

\rowcolor{groupgray}
\multicolumn{6}{l}{
\textbf{Shared AWS Textract OCR}
\hspace{0.4em}\textit{(OCR fixed; detector varies)}
} \\

Presidio-best
& 3,182
& 16.9
& 37.2
& 45.9
& 0.1 \\

GLiNER-base
& 3,052
& 18.9
& 29.0
& 52.1
& 0.0 \\

GLiNER-multi-PII
& 3,204
& 16.3
& 48.8
& 34.9
& 0.1 \\

NVIDIA GLiNER PII
& 3,127
& 18.2
& 22.7
& 59.0
& 0.0 \\

GLiNER2
& 3,244
& 17.3
& 34.8
& 47.9
& 0.0 \\

Amazon Comprehend PII
& 2,499
& 20.3
& 23.0
& 56.5
& 0.2 \\

OpenAI Privacy Filter
& 2,984
& 18.9
& 44.4
& 36.6
& 0.1 \\

\rowcolor{groupgray}
\multicolumn{6}{l}{
\textbf{Vendor-managed OCR}
\hspace{0.4em}\textit{(OCR text unavailable)}
} \\

Google DLP
& 4,127
& --
& 15.5
& 84.5
& 0.0 \\

\hline
\end{tabular}

\vspace{1.2em}

\begin{tabular}{l r r r r}
\hline
\textbf{System} &
\textbf{Leaked critical} &
\textbf{Value unavailable} &
\textbf{Attrib. $n$} &
\textbf{OCR partial} \\
\hline

Presidio
& 3,397
& 26
& 3,371
& 312 \\

Presidio-best
& 3,208
& 26
& 3,182
& 192 \\

GLiNER-base
& 3,077
& 25
& 3,052
& 219 \\

GLiNER-multi-PII
& 3,231
& 27
& 3,204
& 176 \\

NVIDIA GLiNER PII
& 3,152
& 25
& 3,127
& 215 \\

GLiNER2
& 3,271
& 27
& 3,244
& 205 \\

Google DLP
& 4,160
& 33
& 4,127
& -- \\

Amazon Comprehend PII
& 2,518
& 19
& 2,499
& 162 \\

OpenAI Privacy Filter
& 3,008
& 24
& 2,984
& 209 \\

\hline
\end{tabular}

\caption{Attribution of leaked Direct and Linkage identifiers for
OCR-dependent systems using type-agnostic matching at
IoU $\tau=0.75$. The upper panel reports exclusive failure shares;
the lower panel gives their denominators and secondary diagnostic
counts. Value-unavailable instances remain in the leakage total but
are excluded from attribution percentages. OCR partial is a secondary
flag on OCR-miss instances. Dashes indicate unavailable vendor OCR.}
\label{tab:ocr_attribution}
\end{table*}

\begin{table}[H]
\centering
\small
\begin{tabular}{l r r r}
\hline
\textbf{OCR engine} &
\multicolumn{2}{c}{\textbf{Unrecoverable}} &
\textbf{Partial $n$} \\
& $\mathbf{n}$ & $\mathbf{\%\downarrow}$ & \\
\hline
AWS Textract & 623 & \textbf{14.6} & 262 \\
Tesseract    & 765 & 17.9          & 350 \\
\hline
\end{tabular}
\caption{OCR recoverability over 4,280 value-bearing critical
annotations. Of 4,315 critical annotations, 35 lack a verbatim
gold value and are excluded. Partial denotes unrecoverable values
with OCR-window similarity $\geq 0.8$.}
\label{tab:ocr_recoverability}
\end{table}

\end{document}